\documentclass[10pt,twocolumn,letterpaper]{article}

\usepackage{cvpr}              % To produce the CAMERA-READY version
\usepackage{graphicx}
\usepackage{amsmath}
\usepackage{amssymb}
\usepackage{booktabs}
\usepackage{times}
\usepackage{epsfig}
\usepackage{float}
\usepackage{multirow}
\usepackage{enumitem}
\usepackage{graphicx, amsmath, amssymb, caption, subcaption, multirow, overpic, textpos}
\usepackage[table,dvipsnames]{xcolor}
\usepackage[british, english, american]{babel}
\definecolor{citecolor}{HTML}{0071BC}
\definecolor{linkcolor}{HTML}{ED1C24}
\usepackage[colorlinks,linkcolor=blue]{hyperref}
\def\cvprPaperID{**}
\def\confName{****}
\def\confYear{****}

\newlength\savewidth
\newcommand{\tablestyle}[2]{\setlength{\tabcolsep}{#1}\renewcommand{\arraystretch}{#2}\centering\footnotesize}
\renewcommand{\paragraph}[1]{\vspace{1.25mm}\noindent\textbf{#1}}

\newcolumntype{x}[1]{>{\centering\arraybackslash}p{#1pt}}
\newcolumntype{y}[1]{>{\raggedright\arraybackslash}p{#1pt}}
\newcolumntype{z}[1]{>{\raggedleft\arraybackslash}p{#1pt}}

\newcommand{\app}{\raise.17ex\hbox{$\scriptstyle\sim$}}

\definecolor{deemph}{gray}{0.6}

\definecolor{baselinecolor}{gray}{.9}

\begin{document}
	%%%%%%%%% TITLE
	\title{MILES: Visual BERT Pre-training with Injected Language Semantics\\ for Video-text Retrieval}
	
	\author{Yuying Ge$^1$ \quad
		Yixiao Ge$^2$ \quad
		Xihui Liu$^4$ \quad
		Alex Jinpeng Wang$^5$ \\
		Jianping Wu$^6$ \quad
		Ying Shan$^2$ \quad
		Xiaohu Qie$^3$ \quad
		Ping Luo$^1$ \quad\\
		{$^1$The University of Hong Kong} \quad 
		{$^2$ARC Lab, $^3$Tencent PCG}  \quad
		{$^4$UC Berkeley} \\
		{$^5$National University of Singapore} \quad
		{$^6$Tsinghua University} \quad\\
		\tt\small{yuyingge@hku.hk \quad \{yixiaoge, yingsshan, tigerqie\}@tencent.com}\\
		\tt\small{xihui.liu@berkeley.edu \quad jinpengwang@u.nus.edu \quad jianping@cernet.edu.cn \quad pluo@cs.hku.hk}\\
		\vspace{5pt}
	}
	
\maketitle
%%%%%%%%% ABSTRACT
\vspace{-25pt}
	\begin{abstract}
		Dominant pre-training work for video-text retrieval mainly adopt the ``dual-encoder'' architectures to enable efficient retrieval, where two separate encoders are used to contrast global video and text representations, but ignore detailed local semantics. The recent success of image BERT pre-training with masked visual modeling that promotes the learning of local visual context, motivates a possible solution to address the above limitation. In this work, we for the first time investigate masked visual modeling in video-text pre-training with the ``dual-encoder'' architecture. We perform Masked visual modeling with Injected LanguagE Semantics (MILES) by employing an extra snapshot video encoder as an evolving ``tokenizer'' to produce reconstruction targets for masked video patch prediction. Given the corrupted video, the video encoder is trained to recover text-aligned features of the masked patches via reasoning with the visible regions along the spatial and temporal dimensions, which enhances the discriminativeness of local visual features 
	and the fine-grained cross-modality alignment. Our method outperforms state-of-the-art methods for text-to-video retrieval on four datasets with both zero-shot and fine-tune evaluation protocols. Our approach also surpasses the baseline models significantly on zero-shot action recognition, which can be cast as video-to-text retrieval. 
	\end{abstract}

	\section{Introduction}
	Pre-training visual-language models to learn transferable representations for downstream video-text retrieval has attracted increasing attention in recent years.
	The ``dual-encoder'' architectures~\cite{frozen,mil,videoclip,coot,taco,avlnet,support}, where two individual encoders are used to contrast global video and text representations, become the most popular practice to enable efficient retrieval.
	Despite the high efficiency, recent studies~\cite{univl} have shed light on the limitations of such dual-encoder representation learning: the coarse-grained alignment constraint on global video-text features hinders the capture of detailed local semantics and the further improvements in video-text retrieval.
	
	\begin{figure*}[ht]
		\centering
		\includegraphics[width=1.0\linewidth]{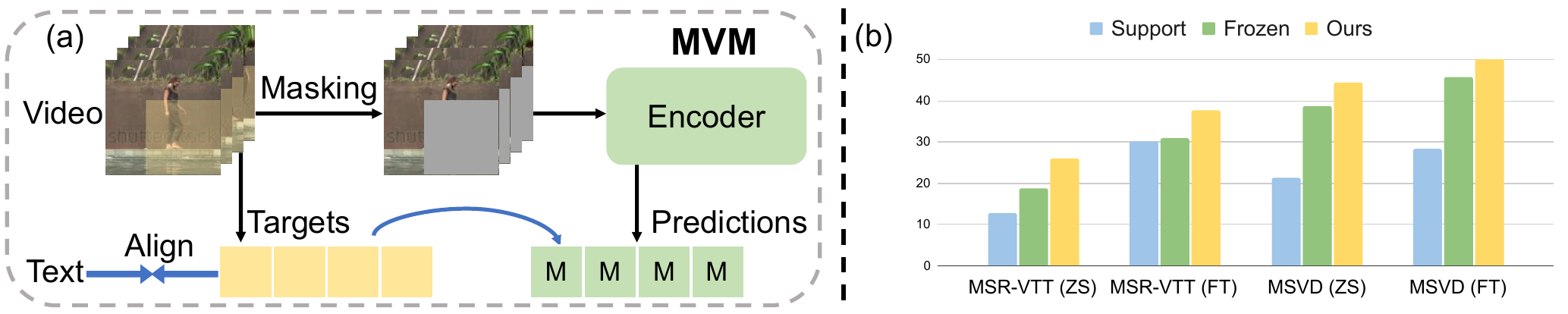}
		\vspace{-15pt}
		\caption{\label{fig:overview} \textbf{(a)} The diagram of \textbf{M}asked visual modeling with \textbf{I}njected \textbf{L}anguag\textbf{E} \textbf{S}emantics (MILES) in video-text pre-training, which aims to reconstruct the masked video content to the targets that are are aligned with the language semantics through reasoning with the spatial and temporal context of the visible video patches. \textbf{(b)} Comparison between recent ``dual-encoder'' methods for video-text pre-training in text-to-video retrieval on MSR-VTT and MSVD with both the zero-shot (ZS) and fine-tune (FT) evaluation protocols (R@1 as the metric).}
		\vspace{-15pt}
	\end{figure*}
	Inspired by masked language modeling~\cite{bert} in natural language processing, a pioneering work, BEIT~\cite{beit}, introduces the pretext task of masked visual modeling (MVM) to promote the learning of local visual context in image pre-training.
	A proportion of image patches are randomly masked, and the vision Transformer~\cite{vit} is trained to recover the vision tokens that are obtained from a pre-learned image ``tokenizer''~\cite{vae} as the reconstruction targets.
	The great success of MVM in image pre-training provides a possible solution to encourage the learning of local visual semantics in video-text retrieval.

	We would like to recall the key factor that makes MVM successful is the denoising training objective, where the design of the masked prediction targets turns out to be the most critical.
	The follow-up works~\cite{zhou2021ibot,dong2021peco} of BEIT confirm this by introducing better image tokenizers that are more aware of the high-level perceptions.
	So \textit{what makes for good targets of masked visual prediction in video-text pre-training?}
	We argue that, towards the goal of accurate video-text retrieval, besides the spatial and temporal visual context understanding, the local alignment with language semantics serves as another important objective for the prediction of masked video patches as illustrated in Fig.~\ref{fig:overview} (a).

	To build masked video prediction targets with injected language semantics, we use a snapshot video encoder as an evolving ``tokenizer'' to produce regressed targets for masked video patches.
	The snapshot video encoder aggregates the knowledge of the in-training video encoder in prior epochs, whose predictions gradually approach the text domain under the global video-text contrastive constraint.
	By imposing MVM regularizations towards the targets from the snapshot encoder, the in-training video encoder can be iteratively improved to capture more detailed video semantics that are locally aligned with the languages, which, in turn, further enhances the evolving ``tokenizer'', \textit{i.e.}, snapshot encoder.
	Our method successfully applies the idea of MVM in video-text pre-training without extra pre-training stages for obtaining a proper cross-modality ``tokenizer''.

	Specifically, based on the ``dual-encoder'' architecture, we employ an extra snapshot video encoder to provide masked vision modeling regularization only for pre-training if not specified, that is, retaining the high efficiency of an ordinary dual-encoder structure in retrieval.
	In each iteration, 
	we randomly mask a large proportion of patches in sparsely sampled videos along both the spatial and temporal dimensions to enforce a high-level understanding of local contents and temporal dynamics.
	The masked videos are fed into the video encoder for performing denoising auto-encoding, while the raw videos are fed into the snapshot video encoder for producing the reconstruction targets.
	Intuitively, given the highly corrupted video, the video encoder is trained to recover text-aligned features of the masked video content via reasoning among the visible regions along the spatial and temporal dimensions, enhancing not only the discriminativeness of local visual features but also the fine-grained cross-modality alignment.

	Our contributions are three-fold.
	(1) We are the first to explore the potential of BERT-style pre-training in video-text retrieval with dual-encoder models. We study the pretext task of masked visual modeling in video-text pre-training and indicate its advantages in both fine-grained video context understanding and video-text local semantic alignment.
	(2) We introduce a flexible and effective method with a snapshot video encoder as the evolving ``tokenizer'' to produce learning targets for the masked video patch prediction. The video encoder gradually improved by the denoising regularizations can be used, in turn, to enhance the ``tokenizer''.
	(3) Extensive empirical results on text-to-video retrieval on four datasets with both zero-shot and fine-tune evaluation protocols fully demonstrate the superiority of our method (Fig.~\ref{fig:overview} (b)). We further evaluate zero-shot action recognition on two datasets, which can be cast as a video-to-text retrieval task. Our method significantly surpasses its competitive counterparts by a large margin. As an additional benefit, we surprisingly find that our method achieves competitive performance on single-modality action recognition with much fewer video hours for pre-training.

	\section{Related Work}
	\noindent\textbf{Pre-training for video-text retrieval.}
	Previous pre-training methods for video-text retrieval can be divided into two categories. 
	Methods in the first category~\cite{frozen,mil,videoclip,taco,coot,avlnet,support,multi,expert,bridgeformer,object}, adopt the ``dual-encoder'' architectures, where 
	two individual encoders are used to contrast global video and text representations, and contrastive learning~\cite{contrastive1,contrastive2} is utilized to distinguish paired video-text data with unpaired data. Although these methods are efficient for video-text retrieval, they ignore local semantics and fine-grained alignment between modalities.
	Methods in the second category~\cite{clipbert,videobert,actbert,hero,univl,vlm} adopt the ``joint-encoder'' architectures to interact cross-modality local features through concatenating videos and texts as inputs with a binary classifier to predict whether videos and texts are aligned or not. Despite they can build local associations between videos and texts, they sacrifice the retrieval efficiency since every text-video pair needs to be fed into the encoder during inference.
	In this work, we adopt the ``dual-encoder'' architecture for efficient retrieval and use the pretext task of masked visual modeling in video-text pre-training to enhance both fine-grained video context understanding and video-text local semantic alignment.
	
	\noindent{\textbf{Image Pre-training with Masked Visual Modeling.}} Recent works introduce masked visual modeling (MVM) to image self-supervised pre-training, where MVM masks a proportion of the visual patches and optimizes the vision Transformers to reconstruct the missing content. The reconstruction targets are proven to be the most critical. 
	For example, \cite{mae} reconstructs the masked image patches in the pixel space, which makes the model focus on short-range dependencies and high-frequency details. \cite{beit} predicts the discrete visual tokens from a pre-learned image ``tokenizer''~\cite{vae}, which requires one more pre-training stage on extra data (250M images~\cite{vae}). 
	Our method uses a snapshot encoder as the evolving ``tokenizer'' without additional pre-training stages.
	A concurrent work, iBOT \cite{zhou2021ibot}, uses a similar online tokenizer to guide MVM for image pre-training. However, we have different purposes. While iBOT encourages MVM to focus more on the high-level visual semantics rather than trivial low-level reconstruction, our method uses a self-training pipeline to progressively inject the text-aware semantics into the MVM targets, aligning the text and visual domains in both global and local representations.
	
	\noindent\textbf{Masked Region/Frame Modeling in Video-text Pre-training.}
	Similar techniques, masked region modeling (MRM) and masked frame modeling (MFM), were introduced in the ``joint-encoder'' methods for video-text pre-training. For example, \cite{actbert} uses MRM, which masks the object regions of video frames with a pre-trained detection model~\cite{faster} and predicts a distribution over fixed vocabulary for the masked-out frame region. \cite{univl,hero,vlm} adopts MFM, which masks video frames and recovers the masked frames to the features encoded from an off-the-shelf video feature extraction network~\cite{s3d}. Both MRM and MFM rely on pre-trained models with extra data to obtain visual-only reconstruction targets, while our work evolves a snapshot video encoder to provide video-text aligned reconstruction targets without additional training stage on extra data.
	
		\begin{figure*}[t]
		\centering
		\includegraphics[width=1.0\linewidth]{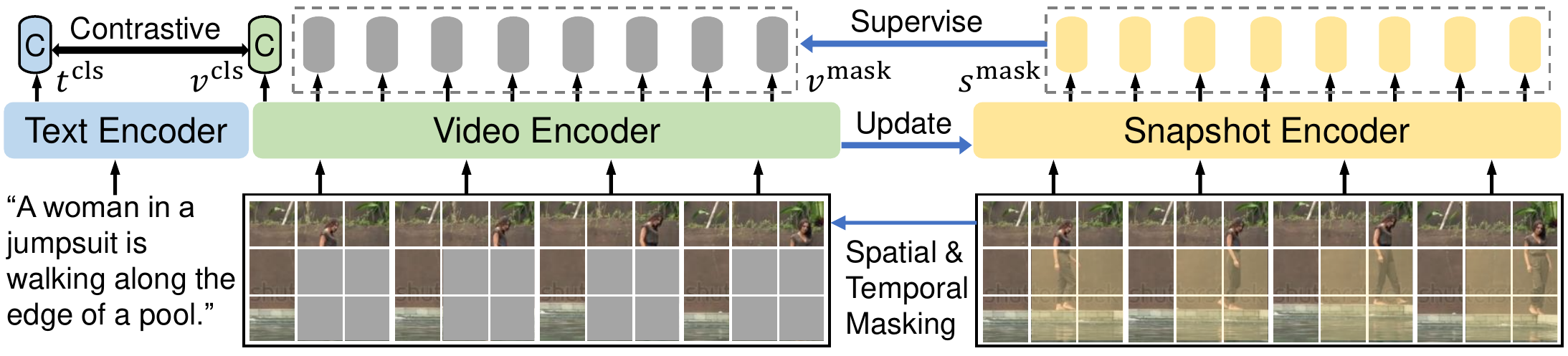}
		\vspace{-15pt}
		\caption{\label{fig:method} The video-text pre-training pipeline of \textbf{M}asked visual modeling with \textbf{I}njected \textbf{L}anguag\textbf{E} \textbf{S}emantics (MILES). Based on the ``dual-encoder'' architecture, \textit{i.e.} a text encoder and a video encoder, we use an extra snapshot video encoder to provide reconstruction targets with injected language semantics for MVM. We first mask a proportion of a video along the spatial and temporal dimension. The masked video is fed into the video encoder to predict features of the masked video patches as $v^{\text{mask}}$, while the raw video is fed into the snapshot encoder to produce the reconstruction targets as $s^{\text{mask}}$. The snapshot encoder is progressively updated from the in-training video encoder under the constraint of contrasting global video representations $v^{\text{cls}}$ and text representations $t^{\text{cls}}$.}
		\vspace{-15pt}
	\end{figure*}
	
	\section{Method}
	In this work, we adopt the ``dual-encoder'' structure for video-text pre-training to realize efficient retrieval, and perform \textbf{M}asked visual modeling with \textbf{I}njected \textbf{L}anguag\textbf{E} \textbf{S}emantics (MILES) by employing an extra snapshot video encoder to provide reconstruction targets.
	In this section, we first revisit the ``dual-encoder'' structure in Sec.~\ref{sec:revisit}. We then introduce our masked visual modeling in Sec.~\ref{sec:mvm} and the pre-training objectives in Sec.~\ref{sec:objective}. Finally, we explain the detailed model architecture including a video encoder, a text encoder, and a snapshot video encoder in Sec.~\ref{sec:architecture}.

	\subsection{Revisiting Dual-encoder for Video-text Pre-training}\label{sec:revisit}
	As shown in Fig.~\ref{fig:method}, we adopt the ``dual-encoder'' structure to maintain high efficiency for video-text retrieval, which contains a text encoder to encode text representations from natural languages, and a video encoder to produce video representations from raw video frames. We first feed a video to the video encoder and the corresponding text description (\textit{e.g.}, ``A woman in a jumpsuit is walking along the edge of a pool'') to the text encoder to embed their respective representations, which are projected to a common feature space as $v^{\text{cls}}$ and $t^{\text{cls}}$ via two separate linear layers. We calculate the similarity between the video and the text by performing the dot product between two projected embeddings $v^{\text{cls}}$ and $t^{\text{cls}}$. We adopt a contrastive objective~\cite{contrastive1,contrastive2} to maximize the similarity between positive video-text pairs in the batch and minimize the similarity between all other negative combinations in the batch. The separate dual encoders ensure the high efficiency for retrieval since only the dot product between video and text representations is calculated during inference.

	\subsection{Masked Visual Modeling in Video-text Pre-training}\label{sec:mvm}
	{\flushleft \bf Overview.} As illustrated in Fig.~\ref{fig:method}, we impose the regularization with injected language semantics for masked visual modeling (MVM) through using a snapshot video encoder to produce reconstruction targets for masked patches in the video, besides dual encoders for efficient retrieval. Specifically, we train the model using the pretext task of MVM with the following steps. 
	
	\textbf{In the first step}, a sparsely sampled video is divided and projected into a sequence of tokens following~\cite{frozen}. A proportion of video tokens are masked out along the spatial and temporal dimensions by being replaced with a [MASK] token, which is a learnable embedding to indicate masked patches. Positional embeddings are further added to the token sequence after [MASK] token replacement as the input token sequence. \textbf{In the second step}, the input token sequence is fed into the video encoder for performing denoising auto-encoding. The video encoder predicts the features of the masked video tokens as $v^{\text{mask}}$ through resorting to visible video patches of spatial and temporal neighbors. \textbf{In the third step}, the raw video is fed into the snapshot video encoder to provide regularization for MVM. Features of those tokens that correspond to masked video patches for the video encoder are obtained as the reconstruction targets $s^{\text{mask}}$. \textbf{In the fourth step}, token-level supervision is imposed on the video encoder through minimizing the ${\ell}_2$ distance between its predicted features $v^{\text{mask}}$ and the reconstruction targets $s^{\text{mask}}$ embedded from the snapshot encoder.
	
	We progressively update the snapshot video encoder from the video encoder under the constraint of contrasting global video representations $v^{\text{cls}}$ from the video encoder and text representations $t^{\text{cls}}$ from the text encoder, which will be explained in the following section.
	Since the snapshot video encoder aggregates the knowledge of the in-training video encoder, its embedded visual features gradually align with language semantics. Imposing the regularization of MVM towards the output of the snapshot encoder iteratively improves video encoder to capture detailed local visual semantics that are aligned with texts, which in turn enhances the snapshot encoder to provide more effective reconstruction targets. Taking the corrupted video as input, the video encoder is trained to recover text-aligned local features of videos through reasoning with the visible patches in the video along the spatial and temporal dimension, which enhances both the discriminativeness of local features and the fine-grained video-text alignment.
	
	{\flushleft \bf Evolving Snapshot Video Encoder.} We use a snapshot video encoder to embed text-aligned video features as reconstruction targets for MVM, which does not require pre-training with extra data. Inspired by previous work~\cite{dino}, we freeze the snapshot encoder over an epoch, and update its parameters in the $k$-th epoch denoted as $\{{\theta}_s\}_{k}$ with Exponentially Moving Averaged (EMA)~\cite{momentum} mechanism as $\{{\theta}_s\}_{k} = \lambda \{{\theta}_s\}_{k-1} + (1 - \lambda)\{{\theta}_v\}_{k-1}$, where $\{{\theta}_v\}_{k-1}$ denotes the parameters of the video encoder at the end of the $(k-1)$-th epoch. The updating mechanism makes the snapshot encoder evolve more smoothly and thus provide more consistent reconstruction targets for MVM. We also explore other update rules for the snapshot encoder in Sec.~\ref{sec:ablation}, which show inferior performances.
	
	{\flushleft \bf Masking Strategy.} Videos usually exhibit similar visual patterns in adjacent patches within a frame or patches of neighboring frames (spatio-temporal neighbors), which makes the masked video patches easy to recover through interpolating between the spatio-temporal neighbors. To make masked visual modeling more challenging and improve model's spatial and temporal understanding, we adopt the ``tube'' masking strategy, which masks blocks of video patches along the spatial and temporal dimension instead of independently masking random patches for each frame. Specifically, we first sample a 2D mask through block-wise masking following~\cite{beit}, and then extend the 2D mask to 3D mask through repeating it in the temporal dimension, such that the spatially masked patches are the same for each frame. Our masking strategy refrains the video encoder from reconstructing the masked video content by extrapolation from adjacent visual patterns, and instead requires actual visual reasoning among visible patches along the spatial and temporal dimension. Other masking strategies are discussed in Sec.~\ref{sec:ablation}, which achieve worse results.
	
	\subsection{Pre-training Objectives}\label{sec:objective}
	We combine two objectives to optimize the entire model in an end-to-end manner including a contrastive objective with Noise-Contrastive Estimation (NCE)~\cite{contrastive1,contrastive2} and a regressive objective with ${\ell}_2$ distance, formulated as below, 
	\begin{small}
		\begin{equation}
			\mathcal{L} = \mathcal{L}_\text{vanilla} + \mathcal{L}_\text{mvm} 
		\end{equation}
	\end{small}
	\begin{small}
		\begin{equation}
			\mathcal{L}_\text{vanilla}= \sum_{i=1}^B NCE(v_i^{\text{cls}},t_i^{\text{cls}}) + \sum_{i=1}^B NCE(t_i^{\text{cls}},v_i^{\text{cls}})
		\end{equation}
	\end{small}
	\begin{small}
		\begin{equation}
			\textbf{NCE}(x_i,y_i)= - \text{log}\frac{\text{exp}(x_i^Ty_i/\tau)}{\sum_{j=1}^B\text{exp}(x_i^Ty_j/\tau)} 
		\end{equation}
	\end{small}
	\begin{small}
		\begin{equation}
			\mathcal{L}_\text{mvm}=\sum_{i=1}^B {||s_i^{\text{mask}} - v_i^{\text{mask}}||}_2
		\end{equation}
	\end{small}	
	where $B$ is the batch size and $\tau$ is the temperature hyper-parameter.
	
	\subsection{Model Architecture}\label{sec:architecture}
	{\flushleft \bf Video Encoder.} Video encoder takes the video as input and produces final video representations and predicts features of the masked video content. A video $V\in R^{M \times 3 \times H \times W}$ sampling $M$ frames is first divided into $M \times N$ patches, and are further fed into a linear projection head to get a sequence of $M\times N$ tokens. We follow BERT~\cite{bert} to add a learnable [CLS] token to the the beginning of the token sequence for global video representations. A proportion of video tokens are replaced with a [MASK] token, which is a learnable embedding. The token sequence after [MASK] token replacement are further added with learnable spatial and temporal positional embeddings. All tokens within the same frame are given the same temporal positional embedding, and all tokens in the same spatial location of different frames are given the same spatial positional embedding, so that the video encoder learns to ascertain the position of video patches. Following \cite{frozen}, the video encoder consists of a stack of space-time self-attention blocks, where each block sequentially performs temporal self-attention and then spatial self-attention on the output of previous block.
	
	{\flushleft \bf Text Encoder.}
	Text encoder takes the nature language as input and outputs final text representations from the [CLS] token, which is concatenated to the beginning of the input text. We adopt a multi-layer bidirectional transformer structure~\cite{distilbert} for the text encoder.
	
	{\flushleft \bf Snapshot Encoder.} Snapshot encoder takes the original video as input and produces the reconstruction targets for MVM only in pre-training and is neglected for retrieval. It has exactly the same architecture as the video encoder.
	
		\begin{table*}\centering
		\caption{Text-to-video retrieval results on MSR-VTT test set, where \textbf{higher} R@k and \textbf{lower} MedR (Median Rank) are better. Zero-shot evaluation results are shown on the top while fine-tuning on the bottom. ``Video Input'' lists the model to extract 3D features, where ``Raw Videos'' means training on raw video frames without pre-extracted features. ``Pairs'' lists the number of video-text pairs for pre-training.} 
		\vspace{-5pt} 
		\scalebox{0.8}{
			\begin{tabular}{c|cccccccc}
				\toprule[1pt]
				\multirow{1}*{Method}&Year&Video Input&Pre-train Dataset&Pairs&R@1&R@5&R@10&MedR\\
				\hline
				\multirow{1}*{ActBERT~\cite{actbert}}&2020&ResNet-3D&HowTo100M&120M&8.6&23.4&33.1&36.0\\
				\multirow{1}*{MMV~\cite{MMV}}&2020&Raw Videos&HowTo100M, AudioSet&138M&9.3&23.0&31.1&38.0\\
				\multirow{1}*{MIL-NCE~\cite{mil}}&2020&Raw Videos&HowTo100M&120M&9.9&24.0&32.4&29.6\\
				\multirow{1}*{VATT~\cite{vatt}}&2021&Raw Videos&HowTo100M, AudioSet&138M&-&-&29.7&49.0\\
				\multirow{1}*{NoiseEst~\cite{noise}}&2021&ResNeXt-101&HowTo100M&110M&8.0&21.3&29.3&33.0\\
				\multirow{1}*{TACo~\cite{taco}}&2021&I3D, S3D&HowTo100M&120M&9.8&25.0&33.4&29.0\\
				\multirow{1}*{VideoCLIP~\cite{videoclip}}&2021&S3D&HowTo100M&110M&10.4&22.2&30.0&-\\
				\multirow{1}*{MCN~\cite{mcn}}&2021&ResNeXt-101&HowTo100M&120M&10.5&25.2&33.8&-\\	
				\multirow{1}*{SupportSet~\cite{support}}&2021&R(2+1)D-34&HowTo100M&120M&12.7&27.5&36.2&24.0\\
				\multirow{1}*{Frozen~\cite{frozen}}&2021&Raw Videos&CC3M, WebVid-2M&5.5M&18.7&39.5&51.6&10.0\\
				\multirow{1}*{AVLnet~\cite{avlnet}}&2021&ResNeXt-101&HowTo100M&120M&19.6&40.8&50.7&9.0\\
				\multirow{1}*{Ours}&2022&Raw Videos&CC3M, WebVid-2M&5.5M&\textbf{26.1}&\textbf{47.2}&\textbf{56.9}&\textbf{7.0}\\
				\toprule[1pt]	
				\multirow{1}*{ActBERT~\cite{actbert}}&2020&ResNet-3D&HowTo100M&120M&16.3&42.8&56.9&10.0\\
				\multirow{1}*{UniVL~\cite{univl}}&2020&S3D&HowTo100M&110M&21.2&49.6&63.1&6.0\\
				\multirow{1}*{MMT~\cite{multi}}&2020&S3D&HowTo100M&120M&26.6&57.1&69.6&4.0\\
				\multirow{1}*{HERO~\cite{hero}}&2021&SlowFast&TV and HowTo100M&120M&16.8&43.4&57.7&-\\	
				\multirow{1}*{NoiseEst~\cite{noise}}&2021&ResNeXt-101&HowTo100M&110M&17.4&41.6&53.6&8.0\\
				\multirow{1}*{ClipBert~\cite{clipbert}}&2021&Raw Videos&COCO, VisGenome&5.6M&22.0&46.8&59.9&6.0\\
				\multirow{1}*{AVLnet~\cite{avlnet}}&2021&ResNeXt-101&HowTo100M&120M&27.1&55.6&66.6&4.0\\
				\multirow{1}*{VLM~\cite{vlm}}&2021&S3D&HowTo100M&110M&28.1&55.5&67.4&4.0\\
				\multirow{1}*{TACo~\cite{taco}}&2021&I3D, S3D&HowTo100M&120M&28.4&57.8&71.2&4.0\\
				\multirow{1}*{SupportSet~\cite{support}}&2021&R(2+1)D-34&HowTo100M&120M&30.1&58.5&69.3&3.0\\	
				\multirow{1}*{VideoCLIP~\cite{videoclip}}&2021&S3D&HowTo100M&110M&30.9&55.4&66.8&-\\
				\multirow{1}*{Frozen~\cite{frozen}}&2021&Raw Videos&CC3M, WebVid-2M&5.5M&31.0&59.5&70.5&3.0\\
				\multirow{1}*{Ours}&2022&Raw Videos&CC3M,  WebVid-2M&5.5M&\textbf{37.7}&\textbf{63.6}&\textbf{73.8}&\textbf{3.0}\\			
				\bottomrule[1pt]
		\end{tabular}}
		\vspace{-10pt}
		\label{tab:msrvtt}
	\end{table*}

	\begin{table*}[t]
	\vspace{1em}
	\centering
	\caption{Experiments of text-to-video retrieval on different datasets, where \textbf{higher} R@k and \textbf{lower} MedR (Median Rank) indicate better performance. We show results with zero-shot evaluation (top) and fine-tuning evaluation (bottom).}
	\vspace{-0.5em}
	\subfloat[
	MSVD test set.
	\label{tab:msvd}
	]{
		\centering
		\begin{minipage}{0.29\linewidth}{\begin{center}
					\tablestyle{1pt}{1.05}
					\begin{tabular}{c|cccc}
						\toprule[1pt]
						\multirow{1}*{Method}&R@1&R@5&R@10&MedR\\
						\hline
						\multirow{1}*{NoiseEst~\cite{noise}}&13.7&35.7&47.7&12.0\\
						\multirow{1}*{SupportSet~\cite{support}}&21.4&46.2&57.7&6.0\\
						\multirow{1}*{Frozen~\cite{frozen}}&38.7&70.1&80.1&2.0\\
						\multirow{1}*{Ours}&\textbf{44.4}&\textbf{76.2}&\textbf{87.0}&\textbf{2.0}\\
						\toprule[1pt]	
						\multirow{1}*{NoiseEst~\cite{noise}}&20.3&49.0&63.3&6.0\\
						\multirow{1}*{SupportSet~\cite{support}}&28.4&60.0&72.9&4.0\\
						\multirow{1}*{Frozen~\cite{frozen}}&45.6&79.8&88.2&2.0\\
						\multirow{1}*{Ours}&\textbf{53.9}&\textbf{83.5}&\textbf{90.2}&\textbf{1.0}\\		
						\bottomrule[1pt]
					\end{tabular}
		\end{center}}\end{minipage}
	}
	\hspace{2em}
	\subfloat[
	LSMDC test set.
	\label{tab:lsmdc}
	]{
		\begin{minipage}{0.29\linewidth}{\begin{center}
					\tablestyle{1pt}{1.05}
					\begin{tabular}{c|cccc}
						\toprule[1pt]
						\multirow{1}*{Method}&R@1&R@5&R@10&MedR\\
						\hline
						\multirow{1}*{AVLnet~\cite{avlnet}}&1.4&5.9&9.4&273.5\\
						\multirow{1}*{NoiseEst~\cite{noise}}&4.2&11.6&17.1&119.0\\
						\multirow{1}*{Frozen~\cite{frozen}}&9.3&22.0&30.1&51.0\\
						\multirow{1}*{Ours}&\textbf{11.1}&\textbf{24.7}&\textbf{30.6}&\textbf{50.7}\\	
						\toprule[1pt]	
						\multirow{1}*{NoiseEst~\cite{noise}}&6.4&19.8&28.4&39.0\\
						\multirow{1}*{MMT~\cite{multi}}&12.9&29.9&40.1&19.3\\
						\multirow{1}*{Frozen~\cite{frozen}}&15.0&30.8&39.8&20.0\\
						\multirow{1}*{Ours}&\textbf{17.8}&\textbf{35.6}&\textbf{44.1}&\textbf{15.5}\\
						\bottomrule[1pt]
					\end{tabular}
		\end{center}}\end{minipage}
	}
	\hspace{2em}
	\subfloat[
	DiDeMo test set.
	\label{tab:didemo}
	]{
		\begin{minipage}{0.29\linewidth}{\begin{center}
					\tablestyle{1pt}{1.05}
					\begin{tabular}{c|cccc}
						\toprule[1pt]
						\multirow{1}*{Method}&R@1&R@5&R@10&MedR\\
						\hline
						\multirow{1}*{VideoCLIP~\cite{videoclip}}&16.6&46.9&-&-\\
						\multirow{1}*{Frozen~\cite{frozen}}&21.1&46.0&56.2&7.0\\
						\multirow{1}*{Ours}&\textbf{27.2}&\textbf{50.3}&\textbf{63.6}&\textbf{5.0}\\
						\toprule[1pt]		
						\multirow{1}*{HERO~\cite{hero}}&2.1&-&11.4&-\\
						\multirow{1}*{CE~\cite{expert}}&16.1&41.1&82.7&8.3\\
						\multirow{1}*{ClipBert~\cite{clipbert}}&20.4&48.0&60.8&6.0\\
						\multirow{1}*{Frozen~\cite{frozen}}&31.0&59.8&72.4&3.0\\
						\multirow{1}*{Ours}&\textbf{36.6}&\textbf{63.9}&\textbf{74.0}&\textbf{3.0}\\
						\bottomrule[1pt]
					\end{tabular}
		\end{center}}\end{minipage}
	}
	%#################################################
	\label{tab:others} \vspace{-.5em}
\end{table*}

	\section{Experiments}
	\subsection{Pre-training Datasets}
	We follow the recent work~\cite{frozen} to jointly pre-train our model on an image dataset Google Conceptual Captions (CC3M)~\cite{cc3m} and a video dataset WebVid-2M~\cite{frozen}. CC3M contains 3.3M image-text pairs with captions harvested from the web. WebVid-2M contains 2.5M video-text pairs with manually generated captions. We do not use the large-scale video-text dataset HowTo100M~\cite{howto100m} with 136M video-text pairs. As \cite{frozen} points out, the captions in HowTo100M are extracted from ASR transcription of continuous narration with incomplete sentences, thus are noisy.
	
	\subsection{Downstream Tasks}
	{\flushleft \bf Text-to-Video Retrieval.} (a). \textbf{MSR-VTT}~\cite{msr} consists of 10K YouTube videos with 200K
	descriptions, which is divided into 9K and 1K videos for training and testing.
	(b). \textbf{MSVD}~\cite{msvd} contains 1,970 videos from YouTube with 80K descriptions. 1200, 100 and 670 videos are split out for training, validation and testing respectively.
	(c). \textbf{LSMDC}~\cite{lsmdc} consists of 118,081 video clips from 202 movies, where the validation set and the test set contain 7,408 and 1,000 videos. %The captions are not concrete enough to describe a specific video, \textit{e.g}. “He nods” is the text description of a video.  
	(d). \textbf{DiDeMo}~\cite{didemo} contains 10K Flickr videos with 40K sentences. 1K videos are split out for testing. We follow \cite{frozen} to concatenate all sentences of a video as a single description.
	We perform evaluation with both the zero-shot and fine-tune setups, and adopt Recall and Median Rank as the evaluation metric.
	
	{\flushleft \bf Action Recognition.}
	(a). \textbf{HMDB51}~\cite{hmdb}, which consists of 6,766 videos with 51 categories.
	(b). \textbf{UCF101}~\cite{ucf}, which consists 13,320 videos with 101 action classes. 
	Both datasets have three standard training/test splits. We explore three experimental settings for evaluation, including \textbf{linear}, where we fix the parameters of the video encoder and only optimize a linear classifier, \textbf{fully fine-tuning}, where we optimize the parameters of the video encoder and the linear classifier together, and \textbf{zero-shot}, where we preform video-to-text retrieval through describing a video with the name of its action class following~\cite{clip}. Averaged results over three training/test splits are reported.
	
	\subsection{Implementation Details}
	For fair comparison, we follow the recent work \cite{frozen} for implementation. We first resize a video to 224 $\times$ 224, and then divide a video into $M$ equal segments. We randomly sample a single frame within each segment for training and uniformly sample a single frame within each segment for testing. We adopt the same model architecture as \cite{frozen} with a video encoder and a text encoder. The video encoder consists of 12 space-time self-attention blocks~\cite{temporal} with patch size $P=16$, and sequence dimension $D=768$. We initialize the video encoder with ViT~\cite{vit} weights trained on ImageNet-21k following~\cite{frozen}. The text encoder is instantiated as DistilBERT~\cite{distilbert} pre-trained on English Wikipedia and Toronto Book Corpus. The snapshot video encoder has the same architecture as the video encoder, and is updated over an epoch with the weights of the video encoder using $\{{\theta}_s\}_{k} = \lambda \{{\theta}_s\}_{k-1} + (1 - \lambda)\{{\theta}_v\}_{k-1}$, where $\theta$ is set as 0.996 in our experiments. We set the dimension of the common feature space as 256 and the temperature hyper-parameter of the contrastive objective as 0.05. For visual augmentation, we randomly crop and horizontally flip during training, and center crop the maximal square crop during testing.
	We adopt a temporal curriculum learning following \cite{frozen}, where we first pre-train our model on the image dataset CC3M and video dataset WebVid-2M sampling 1 frame, and then on the video dataset WebVid-2M sampling 4 frames. Pre-training with 1 frame takes 16 epochs with the batch size of 2048 and the learning rate of $1\times {10}^{-4}$, where the first epoch trains the model with only the contrastive objective as a warm-up. Pre-training with 4 frames takes 4 epochs with the batch size of 1024 and the learning rate of $3\times {10}^{-5}$. We adopt random mask sampling with the masking ratio of 75\% for 1-frame pre-training and block-wise mask sampling with the masking ratio of 75\% for 4-frame pre-training, where the spatial mask is repeated in the temporal dimension. %Pre-training takes a total of 40 hours. 
	For evaluating downstream tasks, we uniformly sample 4 frames for text-to-video retrieval and 16 frames for action recognition following the setting of~\cite{frozen,mil}. We follow \cite{frozen} to expand the temporal embeddings through filling zeros to enable the training of longer frames.

	\subsection{Main Results}
	\subsubsection{Text-to-Video Retrieval}
	Results on MST-VTT~\cite{msr} can be seen in Table.~\ref{tab:msrvtt}. We have the following observations. First of all, our method outperforms all recent work by a large margin. The significant improvement of the performance under the zero-shot evaluation protocol indicates that our pre-trained model is more generalizable that can be used for out-of-domain text-to-video retrieval. Fine-tuning our pre-trained model on the training set of MSR-VTT also achieves better performance.

	Second, the majority of previous work pre-extract 3D features from ``expert'' models as the input of the video encoder (\textit{e.g.} SupportSet~\cite{support} uses expert features from a 34-layer, R(2+1)-D model pre-trained on IG65M~\cite{ig65}). By contrast, our model takes raw video frame pixels as inputs without using any pre-extracted features and surpasses its counterparts.
	Third, previous work mainly pre-train their models on the large-scale HowTo100M~\cite{howto100m}, which is 20$\times$ lager in terms of video-text pairs than CC3M~\cite{cc3m} and WebVid-2M~\cite{frozen}, thus higher computation cost is required. Our pre-trained model achieves higher performance with lower computation cost.
	Finally, some work~\cite{clipbert,hero,univl,vlm,actbert} adopts a joint encoder to concatenate videos and texts as inputs, thus every text-video pair needs to be fed into the encoder during inference, resulting in low efficiency for retrieval. By comparison, our model adopts the efficient ``dual-encoder'' architecture with only a video encoder and a text encoder for inference.

	We further show text-to-video retrieval results on MSVD~\cite{msvd} in Table.~\ref{tab:msvd}, LSMDC in Table.~\ref{tab:lsmdc}, and DiDeMo~\cite{didemo} in Table.~\ref{tab:didemo}. Under both the zero-shot (top) and the fine-tune (bottom) evaluation protocols, our model achieves the best performance on three datasets, demonstrating the effectiveness of our method in utilizing the pretext task of masked visual modeling in video-text pre-training to learn fine-grained cross-modality alignment between videos and texts.
	
	\subsubsection{Action Recognition}
We evaluate zero-shot action recognition on HMDB51~\cite{hmdb} and UCF101~\cite{ucf}, which is cast as video-to-text retrieval by using the name of a video's action class as its description following \cite{clip}. Table.~\ref{tab:zero_action} lists the results, where our model outperforms the competitive baselines by a large margin. Our model improves the averaged top-1 accuracy over three splits by 16.9\% and 10.5\% on HMDB51, 24.9\% and 6.8\% on UCF101 compared with ClipBert~\cite{clipbert} and Frozen~\cite{frozen}. Our method learns powerful cross-modality representations that enable effective zero-shot action recognition.

		\begin{table}\centering
	\caption{Zero-shot action recognition results on HMDB51 and UCF101, with top-1 accuracy as the evaluation metric. ``S'' denotes different testing splits and ``Mean'' reports the averaged results over three splits.} 
	\vspace{-5pt}
	\scalebox{0.75}{
		\begin{tabular}{c|cccc|cccc}
			\toprule[1pt]
			Method&\multicolumn{4}{c|}{HMDB51}&\multicolumn{4}{c}{UCF101}\\
			&S1&S2&S3&Mean&S1&S2&S3&Mean\\
			\hline
			ClipBert~\cite{clipbert}&20.0&22.0&22.3&21.4&27.5&27.0&28.8&27.8\\
			Frozen~\cite{frozen}&27.5&28.3&27.7&27.8&45.4&44.7&47.7&45.9\\
			Ours&\textbf{38.4}&\textbf{38.6}&\textbf{37.8}&\textbf{38.3}&\textbf{51.8}&\textbf{53.4}&\textbf{52.8}&\textbf{52.7}\\	
			\bottomrule[1pt]
	\end{tabular}}
	\vspace{-15pt}
	\label{tab:zero_action}
\end{table}	

	 	\begin{table}\centering
	\vspace{8pt}
	\caption{Comparison of model size and complexity in pre-training and downstream retrieval. ``R@10'' denotes the evaluation results of zero-shot text-to-video retrieval on MSR-VTT.} 
	\vspace{-5pt}
	\scalebox{0.8}{			\begin{tabular}{c|cc|cc|c}
			\toprule[1pt]
			\multirow{2}*{Method}&\multicolumn{2}{c|}{\#params (M)}&\multicolumn{2}{c|}{FLOPs (G)}&\multirow{2}*{R@10}\\
			&train&inference&train&inference&\\
			\hline
			\multirow{1}*{VATT~\cite{vatt}}&414.9&327.0&1004.8&792.0&29.7\\
			\multirow{1}*{Frozen~\cite{frozen}}&180.9&180.9&771.0&771.0&51.6\\
			\multirow{1}*{Ours}&295.1&180.9&1533.4&771.0&56.9\\
			\bottomrule[1pt]
	\end{tabular}}
	\vspace{-15pt}
	\label{tab:flop}
\end{table}

	We further explore the single-modality video representations encoded from our model through evaluating the action recognition under the linear and fine-tuning setups, where the features from the video encoder are extracted and are fed into a linear classifier. As shown in Table.~\ref{tab:action}, our model achieves higher accuracy than some competitive methods, which pre-train their models on datasets with considerably longer video duration. For example, XDC~\cite{XDC} pre-trains on IG-Kinetics that is 14$\times$ longer than our pre-training dataset, but gets worse results.  
Although MMV~\cite{MMV} surpasses our method when it pre-trains the model on HowTo100M and AudioSet (11$\times$ longer than ours) and utilizes extra modalities such as audio and text, its performance is far worse than ours with only text and video or audio and video as inputs. Masked visual modeling with text-aligned features as the reconstruction targets utilizes language semantics more effectively to learn stronger video representations.

	\begin{table}\centering
		\caption{Action recognition results on HMDB51 and UCF101 under linear evaluation (Linear) and fully fine-tuning evaluation (Full), with top-1 accuracy as the evaluation metric. ``Modality'' denotes the extra modality used for pre-training besides videos, \textit{i.e.}, audio (A), text (T), optical flow (OF), motion vector (MV). ``Length'' denotes the video length for pre-training in \textit{K} hours.} 
		\vspace{-5pt}
		\scalebox{0.8}{
			\begin{tabular}{c|cc|cc|cc}
				\toprule[1pt]
				\multirow{1}*{Method}&Modality&Length (K)&\multicolumn{2}{c|}{HMDB}&\multicolumn{2}{c}{UCF101}\\
				&&&Lin&Full&Lin&Full\\
				\hline
				\multirow{1}*{CCL~\cite{CCL}}&-&1.8&29.5&37.8&54.0&69.4\\
				\multirow{1}*{CBT~\cite{CBT}}&-&1.8&29.5&44.5&54.0&79.5\\
				\multirow{1}*{MemDPC~\cite{MemDPC}}&OF&1.8&30.5&54.5&54.1&86.1\\
				\multirow{1}*{CoCLR~\cite{COCLR}}&OF&1.8&52.4&62.9&77.8&90.6\\
				\multirow{1}*{MVCGC}~\cite{compressed}&MV&1.8&53.0&63.4&78.0&90.8\\
				\multirow{1}*{XDC$\_$R~\cite{XDC}}&A&188.3&49.9&61.2&80.7&88.8\\
				\multirow{1}*{XDC$\_$K~\cite{XDC}}&A&188.3&56.0&63.1&85.3&91.5\\
				\multirow{1}*{MIL-NCE~\cite{mil}}&T&134.5&54.8&59.2&83.4&89.1\\
				\multirow{1}*{Frozen~\cite{frozen}}&T&13.0&61.3&66.3&87.8&89.8\\	
				\multirow{1}*{VATT~\cite{vatt}}&A, T&139.8&63.3&-&89.2&-\\
				\multirow{1}*{ELO~\cite{elo}}&A, OF&115.0&64.5&67.4&-&93.8\\
				\multirow{1}*{MMV~\cite{MMV}}&A&134.5&53.6&-&77.1&-\\
				\multirow{1}*{MMV~\cite{MMV}}&T&134.5&55.1&-&86.8&-\\
				\multirow{1}*{MMV~\cite{MMV}}&A, T&139.8&{67.1}&{75.0}&{91.8}&{95.2}\\
				\multirow{1}*{Ours}&T&{13.0}&65.4&70.5&89.5&92.2\\	
				\bottomrule[1pt]
		\end{tabular}}
		\vspace{-15pt}
		\label{tab:action}
	\end{table}

	\subsubsection{Comparing Model Size and Complexity}	
	We analyze the size and the complexity of the model through calculating the number of parameters and FLOPs (higher FLOPs indicate that the model requires more computation costs). As shown in Table.~\ref{tab:flop}, although the snapshot encoder in our method increases the number of parameters and computational costs in pre-training to provide reconstruction targets for MVM, it is not retained for downstream retrieval, rendering an efficient ``dual-encoder architecture with comparable model size and complexity while achieves higher performance.

\begin{figure*}[ht]
	\centering
	\includegraphics[width=0.9\linewidth]{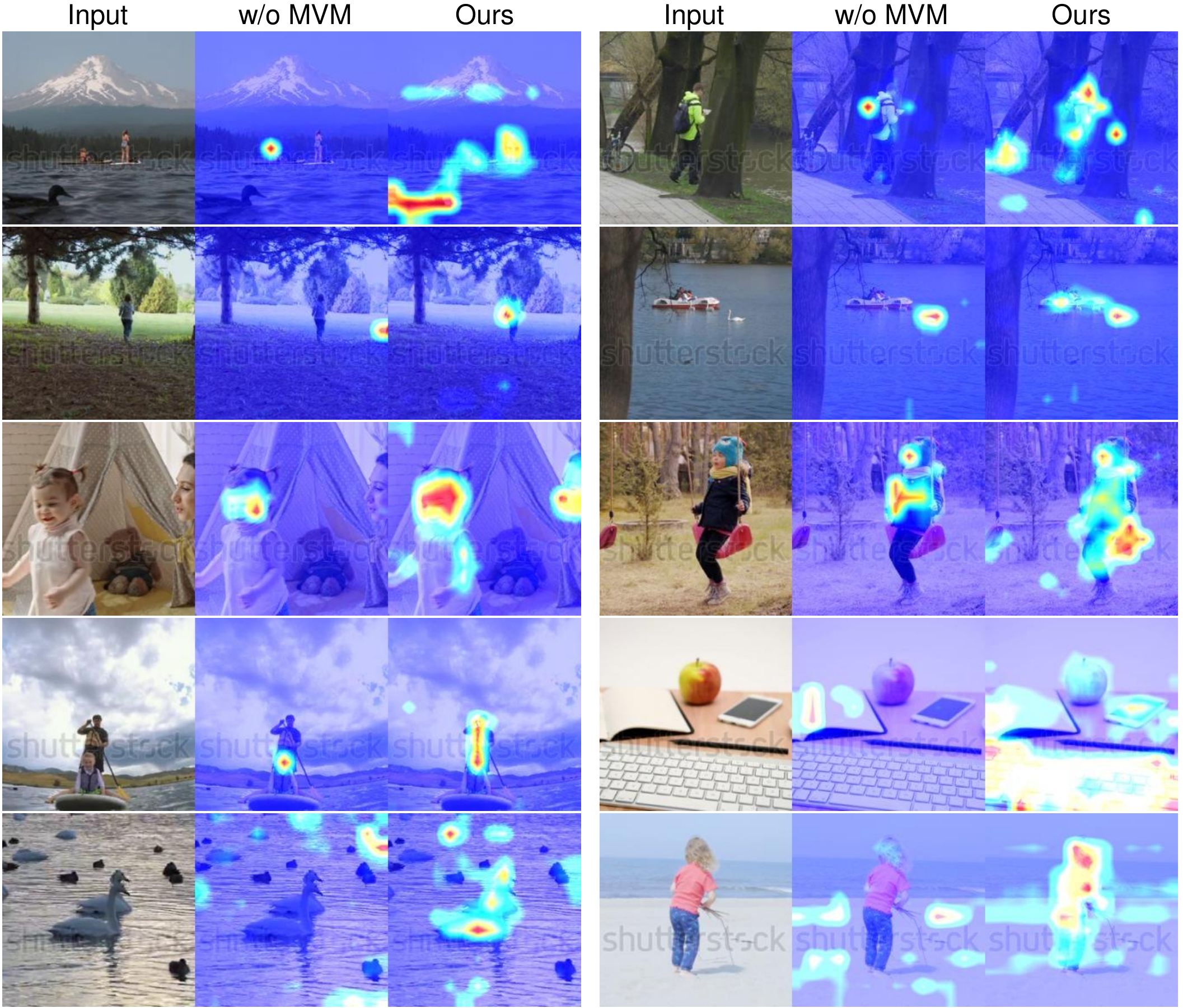}
	\vspace{-5pt}
	\caption{\label{fig:self} The visualizations of the self-attention from the video encoder. Compared with the model that is not trained with MVM, our pre-trained model pays high attention to those significant local regions in the video (\textit{e.g.} a duck in the left column and a bicycle in the right column of the first row), showing that MVM can promote the model to capture fine-grained visual semantics. }
	\vspace{-10pt}
\end{figure*}

	\subsubsection{Visualization}
\noindent\textbf{Local Visual Semantics Capture} We visualize the self-attention map from the video encoder through computing the self-attention of the [CLS] token in the last block. As shown in Fig.~\ref{fig:self}, compared to the model without MVM, our pre-trained model pays high attention to those significant local regions in the video. For example, in the right column of the second row, our model is highly focused on the boat area as well as the duck in the lake while the model without MVM only takes notice of the duck, where the boat region is essential to describe the video content for retrieving this video with the given query text. Pre-training the model with MVM can capture local visual semantics and enhance the fine-grained video context understanding.

\begin{figure*}[ht]
	\centering
	\includegraphics[width=0.9\linewidth]{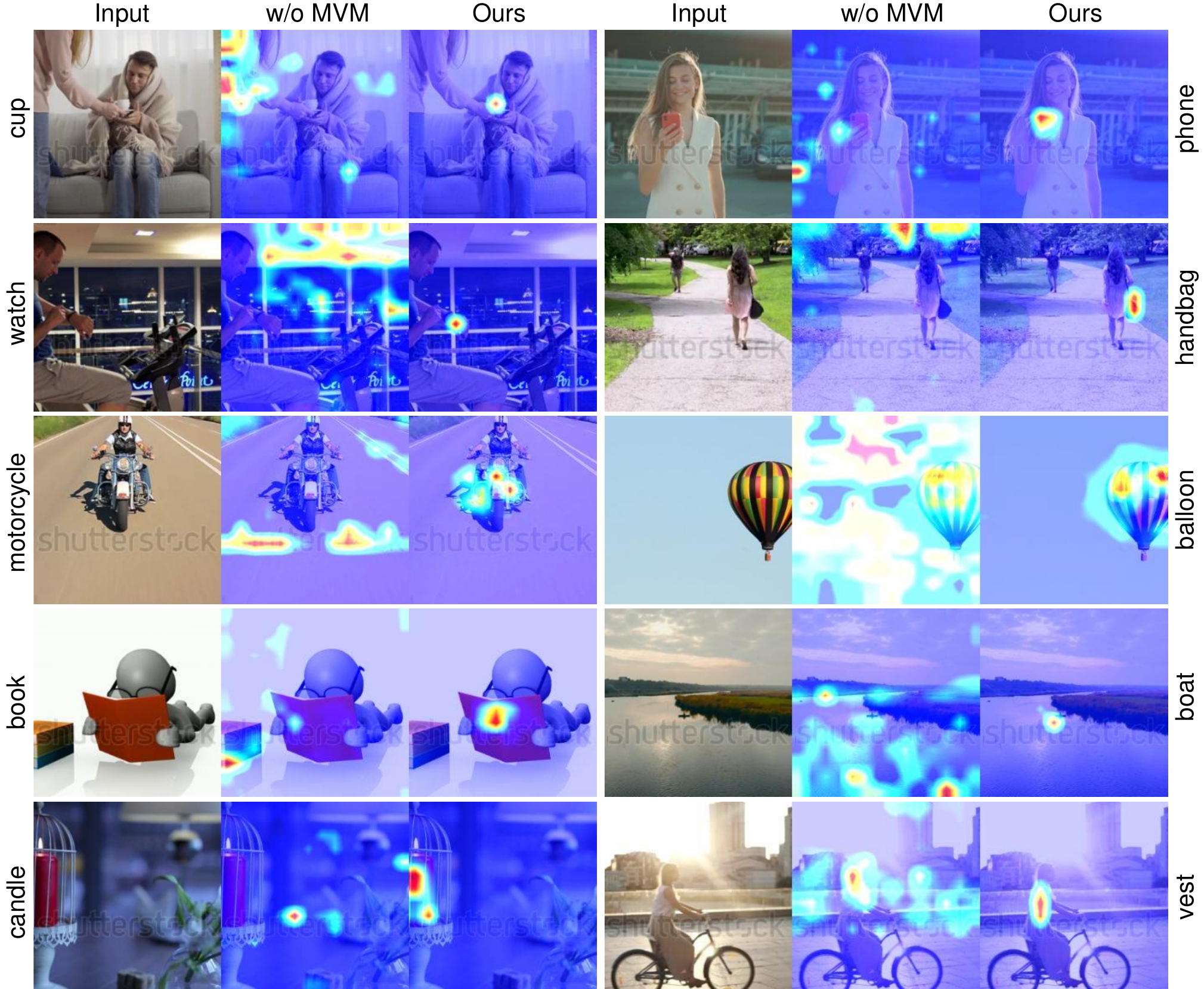}
	\vspace{-5pt}
	\caption{\label{fig:align} The visualizations of the local cross-modality alignment between the text
		token (the words on both sides) and video tokens. Compared with the model that is
		not trained with MVM, our pre-trained model aligns words with corresponding visual
		regions accurately (\textit{e.g.} the region above the man's hand shows large similarity between the word ``cup'' in the left column of the first row), which indicates that the pretext task of MVM can effectively train
		the model to enhance local video-text alignment.}
	\vspace{-5pt}
\end{figure*}

\noindent\textbf{Fine-grained Video-text Alignment} We also visualize the cross-modality alignment between text and video tokens by calculating the similarity map between features embedded from the text encoder and video encoder. Fig.~\ref{fig:align} shows that compared with the model without MVM, our pre-trained model aligns words with corresponding visual regions accurately. For example, in the right column of the first row, visual features of the area where the woman's hand holds the phone shows large similarity with the text features of the word ``phone'' in our model, while the visual features of irrelevant background area are highly similar to the word ``phone'' in the model without MVM. Performing MVM using video-text aligned features as the reconstruction targets effectively trains the model to capture video-text alignment with the ``dual-encoder'' architecture.

\begin{table*}\centering\small
	\vspace{3pt}
	\caption{Text-to-video retrieval results of models initialized from CLIP~\cite{clip} weights on different datasets under zero-shot (top) and fine-tune (bottom) evaluation, where \textbf{higher} R@k and \textbf{lower} MdR (Median Rank) and MnR (Mean Rank) are better.} 
	\vspace{-5pt} 
	\scalebox{0.8}{
		\begin{tabular}{c|ccccc|ccccc|ccccc}
			\toprule[1pt]
			&\multicolumn{5}{c|}{MSR-VTT}&\multicolumn{5}{c|}{MSVD}&\multicolumn{5}{c}{LSMDC}\\
			\multirow{1}*{Method}&R@1&R@5&R@10&MdR&MnR&R@1&R@5&R@10&MdR&MnR&R@1&R@5&R@10&MdR&MnR\\
			\hline
			CLIP-straight~\cite{straight}&31.2&53.7&64.2&4.0&-&37.0&64.1&73.8&3.0&-&11.3&22.7&29.2&56.5&-\\
			CLIP4Clip~\cite{clip4clip}&32.0&57.0&66.9&4.0&34.0&38.5&66.9&76.8&2.0&17.8&15.1&28.5&36.4&28.0&117.0\\
			Ours&\textbf{33.1}&\textbf{59.0}&\textbf{69.9}&\textbf{3.0}&\textbf{25.4}&\textbf{46.8}&\textbf{77.0}&\textbf{85.8}&\textbf{2.0}&\textbf{8.2}&\textbf{15.5}&\textbf{30.3}&\textbf{38.8}&\textbf{23.0}&\textbf{94.8}\\
			\bottomrule[1pt]
			CLIP4Clip~\cite{clip4clip}&43.1&70.4&\textbf{80.8}&2.0&16.2&46.2&76.1&84.6&2.0&10.0&20.7&38.9&47.2&13.0&65.3\\
			Ours&\textbf{44.3}&\textbf{71.1}&80.2&\textbf{2.0}&\textbf{14.7}&\textbf{53.6}&\textbf{81.3}&\textbf{89.9}&\textbf{1.0}&\textbf{5.8}&\textbf{22.5}&\textbf{42.9}&\textbf{50.7}&\textbf{9.5}&\textbf{57.2}\\
			\bottomrule[1pt]
	\end{tabular}}
	\vspace{-10pt}
	\label{tab:clip}
\end{table*}

	\subsubsection{Clip-based Pre-training}
Due to the dominant success of CLIP~\cite{clip} in image-language representation learning, which pre-trains a model with 400 million image-text pairs, recent work~\cite{clip4clip} uses the pre-trained CLIP model as the backbone in video-language pre-training for retrieval. We also pre-train a model initialized from CLIP weights following the setting of \cite{clip4clip} on CC3M and WebVid-2M. Specifically, we use the pre-trained CLIP (ViT-B/32) to initialize the video encoder, text encoder and the snapshot video encoder. As shown in Table.~\ref{tab:clip}, our CLIP-initialized pre-trained model achieves better results for text-to-video retrieval on three datasets with both the zero-shot and fine-tune evaluation protocols. Performing MVM with the video-text aligned features as the reconstruction targets also benefits CLIP-based video-text pre-training for downstream retrieval.

	\subsection{Ablation Study}\label{sec:ablation}
	In this section, we perform ablation studies to discuss the effectiveness of our design on the pretext task of masked visual modeling in video-text pre-training through evaluating different models for zero-shot text-to-video retrieval.
	
	{\flushleft \bf Reconstruction targets.} 
	We explore different reconstruction targets for MVM in video-text pre-training including raw frame pixels as in~\cite{mae}, discrete visual tokens from a learned image ``tokenizer''~\cite{vae} as in~\cite{beit}, and the text-aligned features in this work. As shown in Table.~\ref{tab:target}, using masked visual modeling with different targets improves performance than the baseline model in the first row with only the contrastive objective. Imposing the regularization for MVM with text-aligned features achieves better results than with other visual-only reconstruction targets, which indicates that the local video-text alignment serves as an important objective for the prediction of masked video patches. Besides the ``dual-encoder'' architecture, we further adopt a joint encoder to interact videos with texts for performing denoising auto-encoding, which improves the performance over the ``dual-encoder'' structure with pixels and discrete tokens as the reconstruction targets. Using the joint encoder with the aligned features as the targets brings poorer results, since the learning of cross-modality alignment is mainly achieved by the joint encoder, which does not benefit dual encoders for retrieval. When the reconstruction targets contain video-text aligned semantics, it is more effective to impose MVM regularization on the output local video tokens.

	\begin{table}\centering
	\caption{Ablation studies on reconstruction targets, where ``Joint'' denotes the use of a joint encoder to concatenate videos and texts. Zero-shot text-to-video retrieval on MSR-VTT are evaluated.} 
	\vspace{-5pt}
	\scalebox{0.8}{
		\begin{tabular}{c|c|cccc}
			\toprule[1pt]
			\multirow{1}*{Targets}&Joint&R@1$\uparrow$&R@5$\uparrow$&R@10$\uparrow$&MnR$\downarrow$\\
			\hline 
			\multirow{1}*{-}&$\times$&21.9&43.5&53.5&52.8\\
			\multirow{1}*{Pixels}&$\times$&22.9&44.2&53.7&53.3\\
			\multirow{1}*{Pixels}&$\surd$&24.4&45.2&55.3&53.1\\
			\multirow{1}*{Discrete tokens}&$\times$&24.1&46.6&56.0&51.9\\
			\multirow{1}*{Discrete tokens}&$\surd$&25.0&47.1&56.6&51.8\\
			\multirow{1}*{Aligned features}&$\surd$&25.2&47.1&55.6&48.3\\
			\multirow{1}*{Aligned features}&$\times$&\textbf{26.1}&\textbf{47.2}&\textbf{56.9}&\textbf{46.9}\\
			\bottomrule[1pt]
	\end{tabular}}
	\vspace{-5pt}
	\label{tab:target}
\end{table}

	{\flushleft \bf Updating mechanism of the snapshot encoder.} We compare different strategies to update the snapshot video encode from the video encoder. In our method, we use the video encoder of the previous epoch with momentum update as the snapshot encoder and freeze the the snapshot encoder over an epoch. As shown in Table.~\ref{tab:update}, compared with our updating mechanism in the last row, using the video encoder of the current iteration, or of the previous iteration with or without momentum update all achieve less competitive performance. Since these mechanisms update the snapshot encoder in each iteration, the reconstruction targets are less consistent to impose effective regularization for MVM. Using the video encoder of previous epoch without momentum update also drops performances, since it masks the evolution of the snapshot encoder more sharp. 
	
	\begin{table}\centering
	\vspace{5pt}
	\caption{Ablation studies on the updating mechanism of the snapshot video encoder. ``Mom'' indicates using the momentum update. Zero-shot text-to-video retrieval on MSR-VTT are evaluated.} 
	\vspace{-5pt}
	\scalebox{0.8}{
		\begin{tabular}{cc|cccc}
			\toprule[1pt]
			\multirow{1}*{Mechanism}&Mom&R@1$\uparrow$&R@5$\uparrow$&R@10$\uparrow$&MnR$\downarrow$\\
			\hline 
			\multirow{1}*{Current iter}&-&25.0&47.1&55.7&48.8\\
			\multirow{1}*{Previous iter}&$\times$&24.7&46.4&54.6&48.9\\
			\multirow{1}*{Previous iter}&$\surd$&24.5&46.2&55.3&48.8\\
			\multirow{1}*{Previous epoch}&$\times$&23.8&47.0&56.3&48.2\\
			\multirow{1}*{Previous epoch}&$\surd$&\textbf{26.1}&\textbf{47.2}&\textbf{56.9}&\textbf{46.9}\\
			\bottomrule[1pt]
	\end{tabular}}
	\vspace{-5pt}
	\label{tab:update}
\end{table}

	{\flushleft \bf Masking strategy.} We explore different strategies of masking the videos along the spatial and temporal dimension to perform MVM for multi-frame pre-training. We first propose a frame-wise masking, where a proportion of full frames are masked (\textit{e.g.} 25\% masking ratio means masking one random frame among four sampled frames), so that the model can only resort to neighboring visible frames to recover the masked frames. As shown in Table.~\ref{tab:mask}, frame-wise masking strategy degrades performance, which indicates that reasoning with visible patches along the spatial dimension is also essential. When  ``tube'' masking is not adopted and each frame is spatially masked individually, the model achieves the worst results because the model could temporally ``interpolate'' between frames with visible patches for reconstruction. We further compare different spatial masking strategies with ``tube'' temporal masking. Compared with block-wise masking, which tends to mask large blocks, masking with  random sampling achieves less satisfactory performance because it makes the model easier to reconstruct the masked patches by resorting to visible patches within the frame. Finally, we study the effect of the masking ratio and find that our method with block-wise masking works reasonably well at a ratio of 75\%, but degrades at 65\% when the reconstruction task becomes easier and also degenerates at a ratio 85\% when it is too difficult.

	\begin{table}\centering
	\caption{Ablation studies on the masking strategy. ``Tube'' denotes whether the masked patches are the same for each frame. Zero-shot text-to-video retrieval on MSR-VTT are evaluated.} 
	\vspace{-5pt}
	\scalebox{0.8}{
		\begin{tabular}{ccc|cccc}
			\toprule[1pt]
			\multirow{1}*{Masking}&Ratio&Tube&R@1$\uparrow$&R@5$\uparrow$&R@10$\uparrow$&MnR$\downarrow$\\
			\hline 
			\multirow{1}*{Frame}&25\%&-&25.3&46.2&55.8&47.9\\
			\multirow{1}*{Frame}&50\%&-&25.5&46.6&55.9&47.8\\
			\multirow{1}*{Random}&65\%&$\surd$&24.9&46.6&56.1&47.4\\
			\multirow{1}*{Random}&75\%&$\surd$&25.0&46.8&56.4&47.2\\
			\multirow{1}*{Random}&85\%&$\surd$&25.1&47.0&56.4&47.2\\
			\multirow{1}*{Block}&75\%&$\times$&24.9&45.9&55.4&47.3\\
			\multirow{1}*{Block}&65\%&$\surd$&25.7&46.9&56.2&47.0\\
			\multirow{1}*{Block}&85\%&$\surd$&25.8&46.6&55.5&47.0\\
			\multirow{1}*{Block}&75\%&$\surd$&\textbf{26.1}&\textbf{47.2}&\textbf{56.9}&\textbf{46.9}\\
			\bottomrule[1pt]
	\end{tabular}}
	\vspace{-5pt}
	\label{tab:mask}
\end{table}

		\begin{table}\centering
	\vspace{5pt}
	\caption{Ablation studies on fine-tuning etrieval with MVM. Zero-shot text-to-video retrieval on MSR-VTT are evaluated.} 
	\vspace{-5pt}
	\scalebox{0.8}{
		\begin{tabular}{c|ccccc}
			\toprule[1pt]
			\multirow{1}*{MVM}&R@1$\uparrow$&R@5$\uparrow$&R@10$\uparrow$&R@50$\uparrow$&MnR$\downarrow$\\
			\hline 
			\multirow{1}*{$\times$}&37.7&63.6&73.8&89.8&24.2\\
			\multirow{1}*{$\surd$}&\textbf{37.8}&\textbf{63.6}&\textbf{74.6}&\textbf{90.5}&\textbf{23.9}\\
			\bottomrule[1pt]
	\end{tabular}}
	\vspace{-5pt}
	\label{tab:finetune}
\end{table}

	{\flushleft \bf Fine-tuning with MVM.} When evaluating the model for downstream retrieval with the fine-tune protocol, we only use the contrastive objective to tune our pre-trained model and outperform baselines by a large margin.
	We further use masked visual modeling to fine-tune our pre-trained model on the training set of MSR-VTT and achieve better performance as shown in Table.~\ref{tab:finetune}. We can conclude that MVM is effective to optimize the pre-trained model towards stronger representations with the specific training data.

	\section{Conclusion}
	In this work, we explore masked visual modeling in video-text pre-training with the ``dual-encoder'' architecture for efficient video-text retrieval. We introduce an effective method with a snapshot video encoder to produce reconstruction targets with injected language semantics for the masked video patch prediction, which does not require extra pre-training stages. Training the video encode to recover the text-aligned features of masked video patches through reasoning with the visible regions along the spatial and temporal dimension strengthens both the awareness of local visual features and the fine-grained cross-modality alignment. Extensive evaluations on the text-to-video retrieval and action recognition clearly show the great advantage of our method. 
	
	{\small
		\newpage
		\bibliographystyle{ieee_fullname}
		\bibliography{egbib}
	}
\end{document}